# Creating A New Color Space utilizing PSO and FCM to Perform Skin Detection by using Neural Network and ANFIS

Kobra Nazari[a], Samaneh Mazaheri[b]* and Bahram Sadeghi Bigham[c]

[a] Master's in Computer Science, Department of Computer Science, Vali-e-Asr University of Rafsanjan, Iran kobra.nazari.cs@gmail.com
[b] Assistant Professor, Faculty of Business and Information Technology, Ontario Tech University, Ontario, Canada Samaneh.Mazaheri@ontariotechu.ca
[c] Associate Professor, Department of Computer Science and Information Technology, IASBS, Zanjan, Iran b_sadeghi_b@iasbs.ac.ir

ABSTRACT

Skin color detection is an essential required step in various applications related to computer vision. These applications will include face detection, finding pornographic images in movies and photos, finding ethnicity, age, diagnosis, and so on. Therefore, proposing a proper skin detection method can provide solution to several problems. In this study, first a new color space is created using FCM and PSO algorithms. Then, skin classification has been performed in the new color space utilizing linear and nonlinear modes. Additionally, it has been done in RGB and LAB color spaces by using ANFIS and neural network. Skin detection in RBG color space has been performed using Mahalanobis distance and Euclidean distance algorithms. The BAO database is used for training phase of all the state-of-the-art methods that have been used for comparison. Skin detection's results in new proposed color space utilizing non-linear conversion by ANFIS classification has shown the highest accuracy with 89.22% in compare with other nine recent methods investigated in this paper. In comparison, this method has 18.38% higher accuracy than the most accurate method on the same database. Additionally, this method has achieved 90.05% in equal error rate (1-EER) in testing COMPAQ dataset and 92.93% accuracy in testing Pratheepan dataset, which compared to the previous method on COMPAQ database, 1-EER has increased by %0.87.

1. Introduction

Skin detection is one of the security applications that has been widely used in many areas including face detection in videos [1,2], privacy protection [3], unsupervised hand detection [4], hand gesture detection [5], skin-based user detection [6]. Due to the abundant influence of factors such as race, illumination, different imaging conditions and complex backgrounds with the intensity of various colors in the appearance of the skin, skin detection yet is a complex problem [7].

Currently, there are several research questions that demands answer. For example, is there a less error-prone method to create new color space? If so, do the results show higher speed and accuracy for skin detection in new color space in compare with other color spaces?

The present study used the ideas of Mirghasemi et al. [8] to create a new color space using PSO and FCM. In fact, a new color space will be created with the aim of less error to make skin detection faster and more accurate than RGB and LAB color spaces. Similar database with the one that has been used in Razmjooy et al. [9] will be employed, to demonstrate the expected result and higher accuracy in compare with results in [9]. In fact, performance of new color space is compared with the RGB and LAB color spaces in skin detection. This novel color space can also be used to do image segmentation.

Due to its sensitivity to illumination, RGB color space cannot perform some color processing well [10]. Often due to the integration of data, linear classifiers can hardly classify two classes of skin and non-skin. However, nonlinear classifiers can separate two classes, skin and non-skin, with high accuracy. The purpose of this paper is to introduce a linear and nonlinear conversion of RGB color space through a conversion matrix (W matrix) of size 3×3 to create a new color space. W matrix values are optimized to meet two conditions; firstly, maximizing the distance between centers of skin and non-skin classes, and secondly minimizing entropy of each class.

Several stages of the present paper include first creating an optimal W matrix using PSO and FCM. Secondly, creating a new color space using W matrix of previous step. Thirdly, performing skin detection in new color space in linear and nonlinear states and in RGB and LAB color spaces by utilizing Neural Network and ANFIS classes. Finally, performing skin detection in RGB color space using Mahalanobis and Euclidean distance algorithms. In this study, FNN is implemented and ANFIS is trained with 100 iterations and 15 rules.

This paper begins with Section 2, covering the topic of previous works about skin detection. Section 3, discusses how to create a new color space. Section 4 presents an overview of skin detection method using Neural Network in RGB, LAB and new color space. In Section 5, skin detection using ANFIS will be explained, followed by explaining the skin detection in RGB color space using Euclidean and Mahalanobis distance in Section 6. Next, Section 7 discusses about the databases that have been used in the study. Finally, Section 8 describes experiments, results and concludes this paper with final remarks and future research areas.

2. Related works

There are plenty of related studies using color spaces for skin detection. For example, In 2015, Shaik et al. [11] used the threshold value of three different components in HSV and YcbCr color spaces in dataset including 30 color images to perform skin detection. Experimental result showed that YCbCr is more robust for skin color detection compared to HSV in various illumination condition. In 2017, Song et al. [12] proposed the motion-based skin Region of Interest (ROI) detection method that was performed in a graphics processing unit (GPU)-based a real-time connected component labeling algorithm. They used the RGB histogram to detect skin color range. Their technology is suitable and compatible with natural user interfaces, such as touch less operation systems. In 2017, Brancati et al. [13] evaluated the combinations of chrominance values in the YCb and YCr subspaces of YCbCr to perform skin detection. The proposed method tested in four databases, compared to the six other rule-based methods, it has a better performance evaluation as well as acceptable results even in severe variations in illumination conditions. In 2018, Nguyen-Trang [14] used the Bayesian classifier and connected component algorithm based on RGBUV color space to perform the skin detection. The proposed method is competitive in terms of detection rate and outperformed the others in terms of lower false positive rate and accuracy with a rate of approximately 82%.

Neural networks (NN) enable computer systems to improve their behavior in order to achieve optimal goals based on learning from previous experience [15]. Neural networks are used in many classification issues, such as skin detection. For example, in 2013, Razmjooy et al. [9] proposed a combination of Imperialist Competitive Algorithm (ICA) and Artificial Neural Network (ANN) for skin classification. In their approach, Multi-Layer Perceptron Networks (MLPs) has been employed to solve problem constraints and ICA algorithm is utilized to search for high quality solutions and to decrease the mean square error. Skin detection has been done using RGB color space in linear mode utilizing Baryonic Acoustic Oscillations (BAO) database. Experimental results show that This method can increase the performance improvement of the MLP algorithm and reduced the number of false detections compared to the gradient descent algorithm. In this paper, the highest CDR rate is 70.84% using HNNICA classifier for skin detection. In 2015, Al-Mohair et al. [16] combined a Multilayer Perceptron artificial neural network with K-means clustering method in YIQ color space to perform the skin detection. The proposed extracted the texture and color information to design an efficient skin color classifier. The experimental results showed that proposed algorithm can get great accuracy with an F1-measure of 87.82% using images from the ECU database. In 2016, Gupta and Chaudhary [17] created a system with three robust algorithms, based on RGB, HSV and YCbCr color spaces which are switched according to the statistical mean of the skin pixels. Experimental results showed that neural network technique has made adaptive color space switching to skin classification more accurately compared to the Max Connected matrix. In 2017, Zuo et al. [18] merged the recurrent neural networks (RNNs) layers into the fully convolutional neural networks (FCNs) to perform skin detection. RNN layers improved skin detection in complex backgrounds. Experimental results show that the (FCNs+RNNs) algorithm outperformed standard methods on both the COMPAQ and ECU skin datasets. In experiments, they used 4,670 skin images from the COMPAQ database, and among them 976 were randomly assigned as a test set and the rest as a training set. In COMPAQ database, the highest skin detection performancewas obtained using FCN8S + RNN network, which was 95.93% and 90.18% using area under curve (AUC) and equal error rate (1-EER) metrics, respectively. In ECU database, the highest skin detection performance was obtained using FCN8S + RNN network, which was 98.10% and 94.80% using AUC and 1-EER metrics, respectively. Later on, in 2019, Kızıltaş et al. [19] applied hybrid mode Complex valued neural network (CVNN) and color space conversion to detect skin in UCI Database. In this paper, skin segmentation in the UCI database with 98.60% accuracy has been done with HSV color transformation, which is higher than Nrgb. In 2020, Paracchini et al. [20] tested deep learning in low resolution grayscale face images and in presence of general illumination to perform the skin detection. Three test datasets are used including MUCT dataset, Helen dataset, Complete dataset (merging of MUCT and Helen datasets). Results showed that this method attained the best results on the MUCT dataset. Applications of this method is a significant step in remote Photoplethysmography (rPPG).

Training of ANFIS with appropriate parameters or optimization algorithms is effective in increasing its accuracy and performance for solve problem [21]. Adaptive Network based Fuzzy Inference System (ANFIS) is also widely used in skin detection. In 2008, Lin et al. [22] used an efficient immune-based particle swarm optimization (IPSO) for a neuro-fuzzy classifier to perform skin detection. Competences of PSO improved the mutation mechanism. The experimental results showed that IPSO method with 90.18 in average testing accuracy rate is more efficient than IA and PSO in accuracy rate and convergence speed on more complex CIT database. In 2010, Zaidan et al. [23] utilized fuzzy inference system in RGB color space to perform skin detection. Compared to existing methods, this method has a high skin detection rates of 87% and low false positive. In 2017, Bush et al. [24] used two classifiers including ANFIS and RBFN in RGB color space, to perform the skin detection. Experimental results showed that Radial Basis Function Networks (RBFN) with 94.52% accuracy outperformed ANFIS having 90.10% accuracy for skin detection.

3. Creating a New Color Space

*3.1. Database and Parameters*

The goal in this section is to create the least error W matrix using FCM and PSO. BAO database [25] has been used for algorithm training (see Fig. 1) and images of high intensity with variety of colors have been used to calculate the accuracy of different state-of-the-art methods in addition to proposed method, in two classes including skin and non-skin (see Fig. 2). The target output is expected to convert skin values to white and non-skin values to black in test image (see Fig. 3). The test image is RGB with a dimension of 322 × 1078, here its RGB values are divided into 255 to be normalized and placed between 0 and 1.

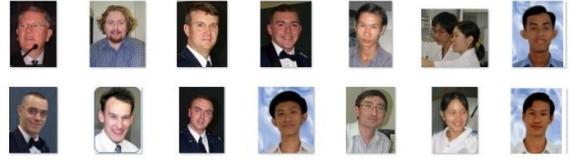

Figure 1. BAO images for training for neural network and ANFIS [25]

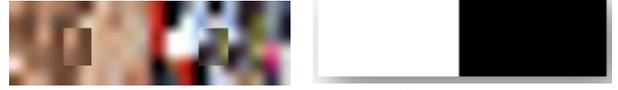

(a)  (b)  Figure 3. Target output
Figure 2. Test set: (a) Skin sets,
(b) Non-skin sets

*3.2 Define PSO Parameters to Create the Conversion Matrix*

Multiple steps of PSO algorithm includes [26]:

1. Define the number of iterations and initial values of velocity and position for the total particle.
2. Calculate Fitness value and particlebest (pbest) values for each particle in each iteration.
3. Calculate the globalbest (gbest) in each iteration.
4. Update the position and speed of the particles.
Repeat steps 1 to 4 to meet the purpose of the problem.

The purpose of the PSO is to produce the W matrix with the least classification error to create optimal color space. In this study, the W matrix, i.e. the position of the particles, is defined as a vector of 1×9 (see Equations 1 and 2). The number of PSO particles is 15, the initial inertia weight is 0.8 and the final initial inertia weight is 0.4, the initial and final values of c1 are 1.75 and 0.5, the initial and final values of c2 are 0.5 and 1.75, respectively. The particles velocity parameter is set to [-0.5, +0.5]. The range of w values is also set to [-5, +5] to search for values at a spatial distance near the target.

Each particle in PSO algorithm has position (i.e. W matrix), velocity and fitness value. For 15 particles, the position vector is randomly generated at [-5, + 5]. The initial velocity vector for all particles is defined as zero, and Fitness value is equal to skin and non-skin classification error (i.e. cost). The goal is to minimize the cost. That is, the lowest cost particle has the most optimal position to create the new color space. For the first particle, the position values (i.e. W matrix) are created as following:

$$W = \begin{bmatrix} 3.1472 & 4.0579 & -3.7301 \\ 4.1338 & 1.3236 & -4.0246 \\ -2.215 & 0.46882 & 4.5751 \end{bmatrix}$$

*3.3. New Color Space Transformations*

Using W matrix described above, test image is transformed from RGB color space to the new color space. Creating a new color space in linear mode is as following:

$$\begin{bmatrix} X_1 \\ X_2 \\ X_3 \end{bmatrix} = \begin{bmatrix} w_{11} & w_{12} & w_{13} \\ w_{21} & w_{22} & w_{23} \\ w_{31} & w_{32} & w_{33} \end{bmatrix} \times \begin{bmatrix} R \\ G \\ B \end{bmatrix} \quad (1)$$

Creating new color space in quadratic mode is as following [8]:

$$\begin{bmatrix} X_1 \\ X_2 \\ X_3 \end{bmatrix} = 0.5 \times (\begin{bmatrix} w_{11} & w_{12} & w_{13} \\ w_{21} & w_{22} & w_{23} \\ w_{31} & w_{32} & w_{33} \end{bmatrix} \times ([R \quad G \quad B] \times \begin{bmatrix} w_{11} & w_{12} & w_{13} \\ w_{21} & w_{22} & w_{23} \\ w_{31} & w_{32} & w_{33} \end{bmatrix})^T)^T$$
(2)

Where the X matrix is the coordinates of each pixel in new color space and the RGB matrix is coordinates of each pixel in the RGB color space. Both linear and non-linear conversion are performed. At this stage, the test image is created in the new color space.

*3.4 Segmentation with FCM*

In this algorithm, first the number of clusters is specified according to the purpose of the problem and then data will allocate to clusters. The allocation criterion is the distance feature. By repeating this process, new centers are obtained by computing the median in each iteration. Following by that, data is allocated to new clusters. This process is repeated until no oscillation is observed in data [27].

Two output classes are defined for FCM algorithm (i.e. skin and non-skin). Once parameter definition is done, options are determined, and test pixels are placed in new color space in FCM input, the algorithm will output two classes with values between 0 and 1. For these values, a threshold is set, that is, where Class 1 is larger than Class 2, and it is inserted in the index vector. Then, the index output is obtained in two classes with values of 0 and 1. After converting the index to the input test image size, a mask-out is obtained. The mask-out image is a logical one where the black values represent non-skin and white values represent skin. In Fig. 4, the FCM output, i.e. mask-out, is shown. For the actual output see Fig. 3.

*3.5. Calculate Fitness Value (Cost)*

A comparison of two logical images (see Fig. 6 and Fig. 7) is performed to calculate the Classification error (i.e. Cost); the error has been calculated 0.15. After calculating the cost and position values for 15 particles, the lowest cost is determined for the first particle as 0.15. It means the minimum classification error (globalbest) for 15 particles is equal to 0.15. This error has been reduced using PSO.

*3.6. Minimizing Cost and Creating Optimal W Matrix by PSO*

In this section, the parameters have been updated. The velocity and position vectors for particles are calculated by Equations (3) and (4), respectively:

$$\text{particle}(i).\text{velocity} = w\_now * \text{particle}(i).\text{velocity} + c_1 * \text{rand}(\text{varsize}) .* (\text{particle}(i).\text{best.position} - \text{particle}(i).\text{position}) + c_2 * \text{rand}(\text{varsize}) .* (\text{globalbest.position} - \text{particle}(i).\text{position}) \quad (3)$$

$$\text{particle}(i).\text{position} = \text{particle}(i).\text{position} + c * \text{particle}(i).\text{velocity} \quad (4)$$

Where parameter c is 1. To accelerate the convergence of the PSO algorithm, the W matrix values range is set to [-5,+ 5] and the particles velocity values range is set to [-0.5,+ 0.5]. In 30 iterations for 15 particles, the values of cost, velocity and particlebest are calculated. The diagrams of cost variations in linear and nonlinear states are shown in Fig. 5. Finally, the globalbest value (i.e. minimum classification error) of 0.11 is obtained. The optimal W matrix corresponding to this error is as following:

$$W = \begin{bmatrix} -3.0403 & 3.341 & 2.9736 \\ 1.4881 & -0.6337 & 2.8407 \\ -1.3716 & -3.8494 & -4.4728 \end{bmatrix}$$

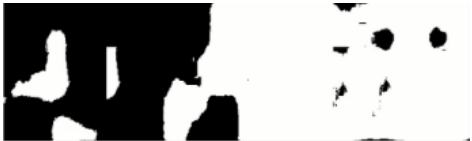
Figure 4. Mask-out (FCM output)

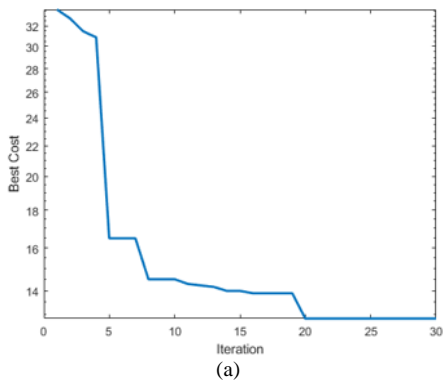
(a)

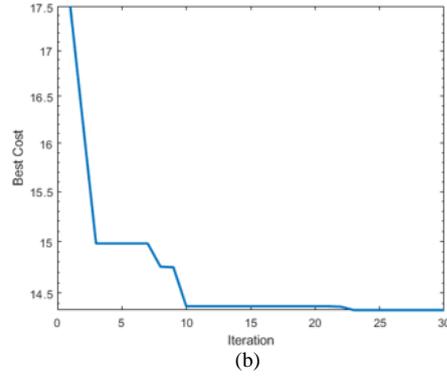
(b)
Figure 5. Charts of cost in linear and nonlinear states. (a) Fitness_L, (b) Fitness_N.

*3.7. Convert images using the W matrix to a new color space*

At this stage, the pixels values of images 4 and 5 are normalized to the range of 0 and 1. Then, using the optimal W matrix (achieved in the previous section) and Equations (1) and (2), are transformed into new color space in linear and nonlinear states. In the next step, skin detection in the above images will be shown using new color space, neural network and ANFIS.

4. Skin detection using Neural Network in RGB, LAB and the New Color Space

1. FNNs are widely used in classification and pattern recognition as well as training [28]. In a Feedforward Multilayer perceptron network, all the neurons (or nodes) in the layers are graphically connected. Each neuron has an activity function and a sum function and communicates with other neurons through its weight. The sum function calculates the product of inputs and weights [29].

In this experiment, Fig. 1 has been used for training and Fig. 2 for neural network and ANFIS testing. The steps of skin detection using the Feedforward neural network in three color spaces has been described in the following sections 4.1 to 4.7.

*4.1. Color Space Selection*

Firstly, there is a need to sample pixels from the image, so they can be categorized into two classes, skin, and non-skin class. 216 pixels have been sampled from Fig. 1 and classified into two classes of skin and non-skin. To detect skin in RGB color space, the R, G and B values of sampled pixels have been divided by 255 to normalize and get a value between 0 and 1. To detect skin in LAB color space, RGB color space has been converted to LAB. Since in the LAB space, the values of L, A, and B of pixels are between 0 and 128, therefore the R, G and B values of the sampled pixels have been divided by 128 to normalize and get a value between 0 and 1. To detect the skin in the new color space, the conversion of RGB color space to the new color space has been performed using the optimal W matrix (obtained above) as well as Equations 1 and 2. In this way the sampled pixels have been placed in the new color space.

*4.2. Create Neural Network Input and Output Vectors*

Normalized pixel values from three color spaces in the vector state of 216×3 have been placed in Train-inputs separately. At Train-targets, the skin and non-skin pixel values have been assigned 1 and 0 labels, respectively.

*4.3. Create Neural Network Input and Output Vectors*

The number of first layer neurons by trial and error is 18, the number of last layer neurons is 1 (equal to the number of output classes i.e. skin class), the Mean Square Error (MSE) is $10^{-10}$, and the number of the times that algorithm repeats is 1000. The first and second layers are sigmoid activator function.

*4.4. Neural Network Training*

The training process will stop at 145 repeats. The MSE value is $10^{-9}$, which is close to the target value.

*4.5. Neural Network Testing by Categorizing Images*

Images in Fig. 1 have been transformed from RGB color space into new and LAB color space. Then, values of all three color spaces have been normalized to a value between 0 and 1, which results in reshaping the size of the input image visually. Following by that, layers R, G and B of images have been extracted and after reshaping, obtained as im_3xn matrices, where n is the number of pixels (similar to Train-inputs, where the pixels values have been in columns). Then, values of im_3xn have been individually converted to 0 and 1 according to the selected color space (RGB, LAB and the new color space). The outputs of this section and the training have been placed at the beginning of the neural network for simulation.

*4.6. Neural Network Simulation*

The values of Train-outputs in each of three color spaces are values between 0 and 1, including zero and one. Since outputs in Train-targets are set to only values 0 and 1, then the values of Train-outputs with a threshold of 0.5 are converted to zero and 1. These values are then converted to their corresponding input image size and reshaped to image mode. The outputs of this section are NN simulation, a logical image in which white areas are detected as skin and black areas are detected as non-skin.

*4.7. Post Processing and Skin Detection*

In NN simulation by defining the disk structural element of size 4, close and open operations are performed to fill holes. By multiplying the images of this step in the corresponding input test image, the skin values have been extracted from each image. In this section, outputs are shown only in the test image. NN simulation for each color space is shown in part b in Figs. 6 to 9 and the final outputs of skin detection are shown in part c of these figures.

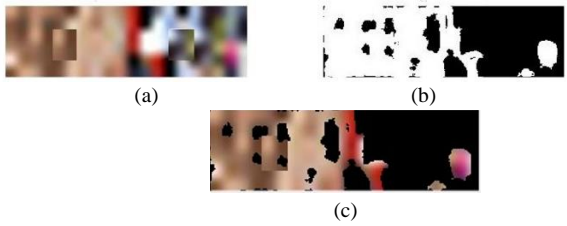

Figure 6. Skin detection in RGB color space using neural network (RGB_NN). (a) Image test in RGB, (b) NN simulation, (c) Skin detection.

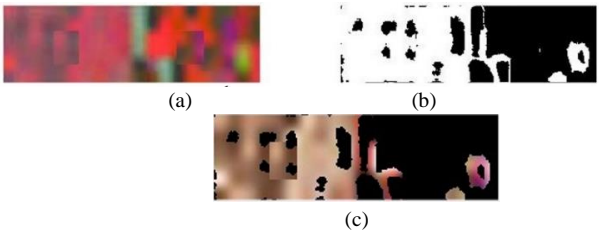

Figure 7. Skin detection in LAB color space using neural network (LAB_NN). (a) Image test in LAB, (b) NN simulation, (c) Skin detection.

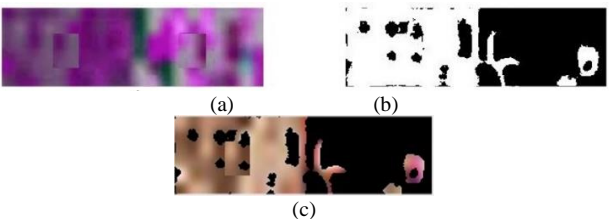

Figure 8. Skin detection by linear transformation in new color space by neural network (L_New_NN). (a) Image test in new color space, (b) NN simulation, (c) Skin detection.

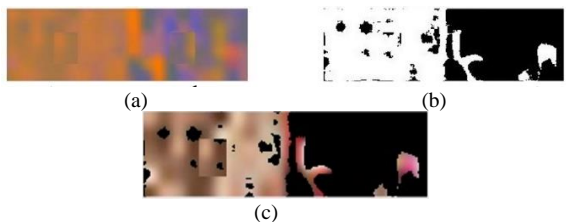

Figure 9. Skin detection by non-linear transformation in new color space by neural network (N_New_NN). (a) Image test in new color space, (b) NN simulation, (c) Skin detection.

*5. Skin Detection utilizing ANFIS*

The abundant use of ANFIS in simulation systems that have high complexity can be observed [30]. The ANFIS structure consists of 5 layers, by passing the trained data from these layers, create a Sugeno fuzzy network. In these nonlinear Fuzzy inference systems, the input-output relation is created with a set of fuzzy rules so as to create the target output. Adaptive technique is a good way to learn data and produce optimal and good results. In this structure, membership functions can be developed either using back propagation or forward propagation systems [31]. The steps of skin detection using ANFIS in three color spaces are described as following.

*5.1. Color space selection*

The steps of sampling and normalizing pixels values are quite similar to the neural network process.

*5.2. Create ANFIS input and output vectors*

Normalized pixels values from all three color spaces in the vector state of $3 \times 216$ have been placed in Train-inputs separately. At Train-targets, values of skin and non-skin pixels have been assigned 1 and 0 labels, respectively. Following by that, the Train-inputs and Train-targets vectors have been placed in Train-data.

*5.3. Define parameters and create Sugeno network*

Exponent value, maximum number of iterations and Maxim-provement value is set to 2, 100 and 1e-2, respectively. These parameters have been placed in the FCM_option input. The number of rules to achieve the goal by trial and error is set to 15. By adding FCM_option values, Train-inputs, Train-targets and the number of rules, the Sugeno network is created and then trained.

*5.4. Parameters definition and ANFIS training*

The number of iterations and the target error is set to 1000 and zero, respectively. The step parameters are also defined. Hybrid method is used for optimization. These parameters and options, Train-data and Sugeno values have been placed in the ANFIS training entry. Finally, the Root Mean Square Error (RMSE) value 0.008 is achieved.

*5.5. ANFIS Testing by Categorizing Images*

This step is very similar to the neural network. At the end of this stage, values of im_3xn have been individually converted to 0 and 1 according to the selected color space (RGB, LAB and new color space).

*5.6. ANFIS Simulation*

In the ANFIS simulation, all the ANFIS Training and im_n×3 values (in rows) have been placed. The values of Train-outputs in each of three color spaces are values between 0 and 1 or less than zero and more than one. Since the outputs in Train-targets are set to only values 0 and 1, then values of Train-outputs with a threshold of 0.5 are converted to zero and 1. These values are then converted to their corresponding input image size and reshaped to image mode. The outputs of this section is ANFIS simulation, a logical image in which white areas are detected as skin and black areas are detected as non-skin.

*5.7. Post Processing and Skin Detection*

Similar to the neural network, morphological processing has been performed on ANFIS simulation images and thus skin values are extracted from images. In this section, outputs are shown only in the test image. ANFIS simulation is shown in part b in Figs. 10 to 13 for each color space. The final output of skin detection is shown in part c in Fig. 10 to 13.

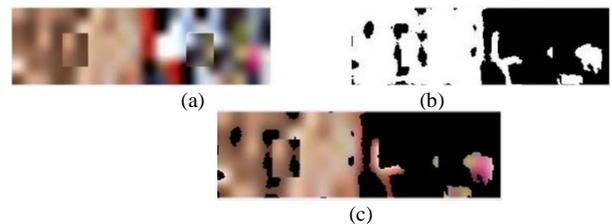

Figure 10. Skin detection in RGB color space using ANFIS (RGB_ANFIS). (a) Image test in RGB, (b) ANFIS simulation and (c) Skin detection.

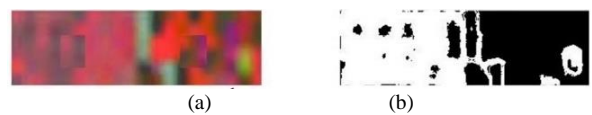

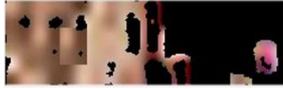
(c)

Figure 11. Skin detection in LAB color space using ANFIS (LAB_ ANFIS). (a) Image test in LAB, (b) ANFIS simulation and (c) Skin detection.

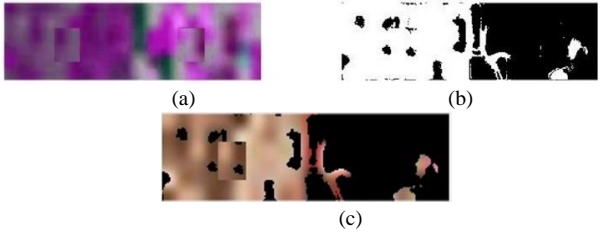

Figure 12. Skin detection by linear transformation in new color space by ANFIS (L_New_ ANFIS). (a) Image test in new color space, (b) ANFIS simulation and (c) Skin detection.

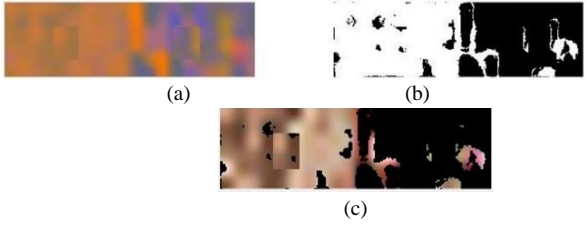

Figure 13. Skin detection by nonlinear transformation in new color space by ANFIS (N_New_ANFIS). (a) Image test in new color space, (b) ANFIS simulation and (c) Skin detection.

6. Skin Detection in RGB using Euclidean and Mahalanobis distance

In this section, the goal is to segment pixels of an RGB image into two classes. The first class will be pixels that are in the specified color range and the second class are the pixels that are not in the specified color range. To this end, color similarity of the pixels must be measured. One method is to calculate Euclidean distance [32]. Here, the output of simulations is shown only in the test image and in part b in Fig. 14 and 15. Similar to the neural network and ANFIS, morphological processing is performed and then skin values have been extracted from the images.

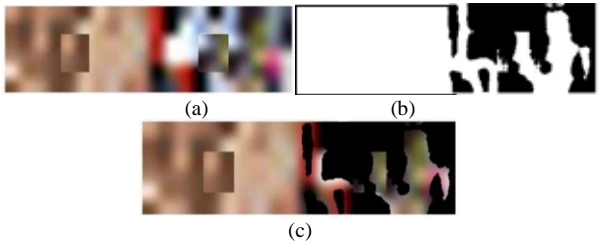

Figure 14. Skin detection in RGB color space using Mahalanobis distance method (RGB_ Mahalanobis). (a) Image test, (b) Mahalanobis simulation and (c) Skin detection.

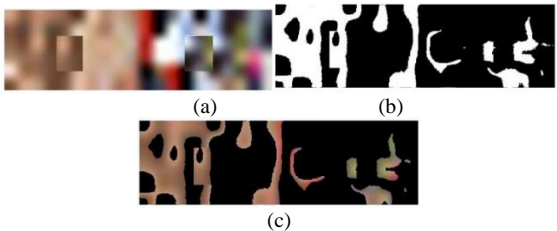

Figure 15. Skin detection in RGB color space using Euclidean distance method (RGB_ Euclidean). (a) Image test, (b) Euclidean simulation and (c) Skin detection.

7. Databases

The following databases are used for training and testing in this paper.

1. BAO dataset: The BAO database containing 370 face images of different races, mostly from Asian, with a wide range of size, background and lighting. 216 skin and nonskin pixels from 14 images of the BAO database have been used to train neural networks and Anfis. (see Figure 1)

2. Test image (Figure 2)

3. COMPAQ database: The COMPAQ dataset, was released by Jones and Regh [33] in 1999, contains 13,634 images (8,964 non-skin images and 4,670 skin images). 976 skin images of this database have been used to test our proposed method.

4. Pratheepan database: It has 78 images, of which 39 of these images have been used to test the proposed method [34].

8. Experiments and Results

To calculate accuracy, NN simulation, simulation ANFIS, simulation Mahalanobis and simulation Euclidean in three color spaces (part b in Fig. 6 to Fig. 15), have been compared separately with the target output (see Fig. 6). Accuracy criteria have been calculated by Equations (5), (6) and (7) [9]. The results of calculating the accuracy and the time taken of the 10 methods are shown in Table 1.

$$CDR = \frac{\text{No. of Pixels Correctly Classified}}{\text{Total Pixels in the Test Dataset}} \quad (5)$$

$$FRR = \frac{\text{No. of Skin Pixels Classified as Non-Skin Pixels}}{\text{Total Pixels in the Test Dataset}} \quad (6)$$

$$FAR = \frac{\text{No. of Non-Skin Pixels Classified as Skin Pixels}}{\text{Total Pixels in the Test Dataset}} \quad (7)$$

Table 1: Accuracy and the time taken of 10 methods reviewed in the proposed skin detection method (unit:%).

| Algorithm | Training images | Testing images | CDR | FRR | FAR | Elapsed time (s) |
|---|---|---|---|---|---|---|
| RGB_NN | 216 pixels from BAO dataset | Figure 2 | 82.83 | 7.01 | 10.15 | 3.003 |
| LAB_NN | | | 85.45 | 6.36 | 8.17 | 5.237 |
| L_New_NN | | | 86.09 | 6.64 | 7.26 | 3.988 |
| N_New_NN | | | 88.28 | 4.53 | 7.18 | 3.970 |
| RGB_ANFIS | | | 86.60 | 5.11 | 8.28 | 4.874 |
| LAB_ANFIS | | | 86.83 | 4.82 | 8.33 | 7.864 |
| L_New_ANFIS | | | 88.68 | 4.95 | 6.36 | 4.901 |
| N_New_ANFIS | | | 89.22 | 3.28 | 7.48 | 4.001 |
| RGB_Mahalanobis | | | 79.22 | 0 | 20.77 | 9.223 |
| RGB_Euclidean | | | 64.49 | 27.48 | 8.02 | |

Table 2: Comparison of the performance of other methods of skin detection with the method proposed (unit:%).

| Algorithm | Training images | Testing images | CDR | FRR | FAR |
|---|---|---|---|---|---|
| HNNICA (Razmjooy et al. [9]) | BAO dataset | Figure 2 | 70.84 | 4.16 | 25 |
| MLP (Razmjooy et al. [9]) | BAO dataset | | 68.42 | 5.2 | 26.38 |
| proposed method (N_New_ANFIS) | 216 pixels from BAO dataset | | 89.22 | 3.28 | 7.48 |

So far, different methods and databases have been used for skin detection. It is difficult to make a fair comparison between the different methods of skin detection due to the lack of several items including a global standard for evaluation of methods, access to other people's databases, quality of some databases [35] and having the same dataset for raining and testing. Razmjooy et al. [9] used BAO database to show the performance as well as calculating the accuracy of skin detection using figure 2 (Test set). Their performance results are shown in Table 1. The method proposed and discussed in this paper has 89.22% accuracy for skin detection in new color space with nonlinear conversion by ANFIS. This value has the highest accuracy in both Tables 1 and 2. The time taken for this method according to Table 1 compared to other methods is desirable. Additionally, skin detection in RGB color space by neural network clustering with an accuracy of 82.83 % is 11.99% more accurate than previous method which is similar in terms of color space and database [9] (see Table 1 and Table 2).

The methods examined in Table 1 were tested with 10 random images in each of the COMPAQ and Pratheepan databases. The results showed that Skin detection's results in new proposed color space utilizing nonlinear conversion by ANFIS classification in accuracy, AUC and 1-EER is higher than 9 the method discussed in this article (see Table 1). Therefore, this method was used to test a larger number of images in two databases.

The results of skin detection in new proposed color space utilizing nonlinear conversion by ANFIS classification on the COMPAQ and Pratheepan databases have been runned 5 times and its best accuraces is shown in Tables 3 and 4, respectively Also, Comparison of proposed method with other skin detection methods on these two databases is shown in tables 5 and 6 . Also, the simulated output of the proposed method is shown in several test images of these two databases in Figures 16 and 17.

Table 3: The results of testing skin detection using proposed method (N_New_ANFIS) on COMPAQ database (unit:%).

| Method | COMPAQ dataset | | | | |
|---|---|---|---|---|---|
| proposed method (N_New_ANFIS) | Training images | Testing images | CDR | FRR | FAR |
| | 216 pixels from BAO dataset | 976 images | 94.79 | 2.63 | 2.56 |
| | | | R 71.23 | F 65.05 | P 61.85 |
| | | | FPR 2.83 | FNR 28.76 | TNR 97.17 |
| | | | TDE 31.59 | Acc 94.79 | AUC 92.99 |
| | | | 1-EER 91.05 | RMSE 0.051 | |

Table 4: The results of testing skin detection using proposed method (N_New_ANFIS) on Pratheepan database (unit:%).

| Method | Pratheepan dataset | | | | |
|---|---|---|---|---|---|
| proposed method (N_New_ANFIS) | Training images | Testing images | CDR | FRR | FAR |
| | 216 pixels from BAO dataset | 39 images | ٩٢,٩٣ | 2.55 | ٤,٥٢ |
| | | | R ٩١,٤ | F ٨٨,٩٩ | P ٨٧,٢٦ |
| | | | FPR ٧,١٢ | FNR ٨,٥٩ | TNR ٩٢,٨٨ |
| | | | TDE ١٥,٧٢ | Acc ٩٢,٩٣ | AUC 95.06 |
| | | | 1-EER 92.25 | RMSE 0.055 | |

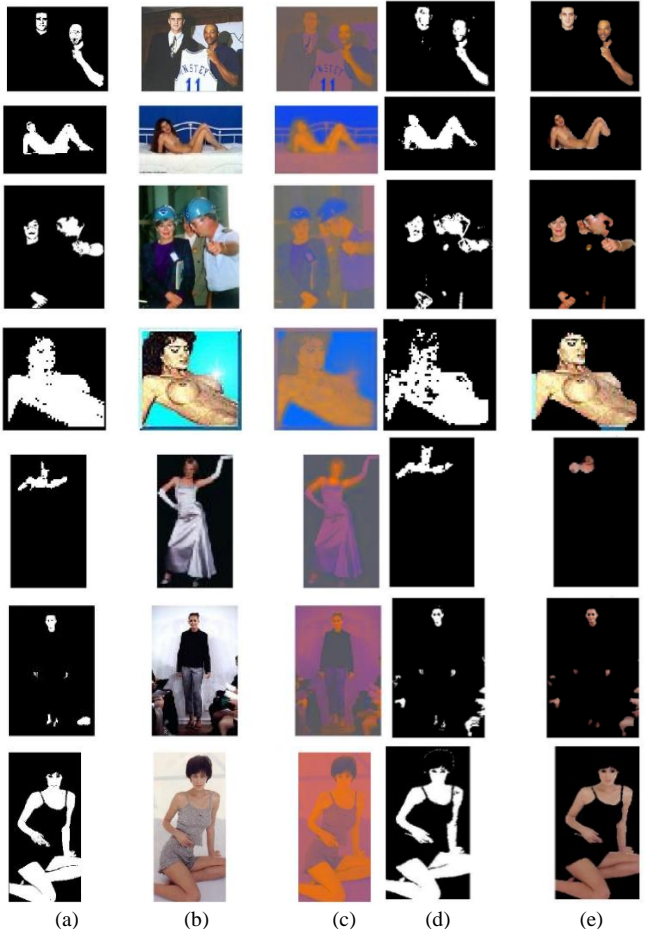

(a) (b) (c) (d) (e)

Figure 16. Examples of Skin detection by nonlinear transformation in new color space by ANFIS (N_New_ANFIS) with the COMPAQ dataset. (a) Orginal Mask, (b) Image test, (c) Image test in new color space, (d) ANFIS simulation and (e) Skin detection.

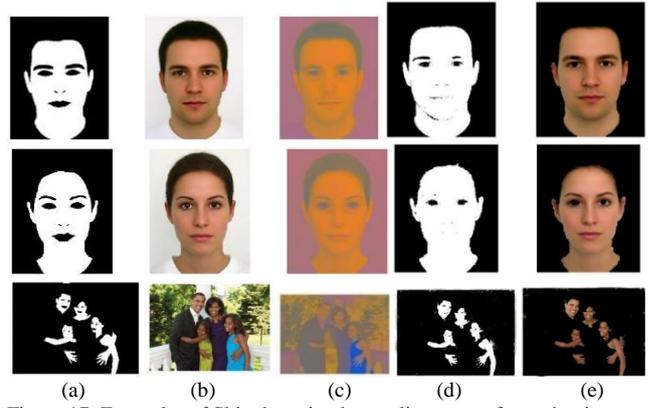

(a) (b) (c) (d) (e)

Figure 17. Examples of Skin detection by nonlinear transformation in new color space by ANFIS (N_New_ANFIS) with the Pratheepan dataset. (a) Orginal Mask, (b) Image test, (c) Image test in new color space, (d) ANFIS simulation and (e) Skin detection.

To evaluate the performance of the proposed skin detection method in the COMPAQ database, several methods from previous advanced work have been cited, such as: Jones and Rehg [33], Sun [36] and Zuo et.al [18] (see Table 5). In order to compare proposed method with these methods, we calculated respective area under curve (AUC) and the receiver operating characteristics (ROC) and equal error rate (1-EER) according to Zuo et al. [18].

Zuo et al. [18] used 3694 images from COMPAQ database for network training and it is a powerful deep learning method. According to Table 5, the AUC value of our proposed method is less than Zuo et al. [18], but in spite of the fact that the number of our training pixels is much less than Zuo et al. [18], and the number of our test images is equal to Zuo et al. [18] (976 images), 1-EER value in our method is higher than other methods in Table 5. The test time of each image is between 3-4 seconds. In some COMPAQ database images, areas of the skin in the orginal mask are labeled black, so accuracy is reduced when comparing these masks to ANFIS simulation (see see Fig. 16). In our method the ROC curve is created by plotting the true positive rate against the false positive rate at various threshold settings [18]. In figure 18 ROC curves of proposed skin detection method has been shown on the COMPAQ and Pratheepan datasets. According to Figure 18 (a and b), the curves at the top left of the ROC space have better performance which have lower FP rate and higher TP rate, also, low values of RMSE in two databases indicate the efficiency of the proposed method. I our method, Due to the low quality and noise and gif images in the COMPAQ database, filters were applied to soften, remove noise and fill holes in the test images. According to Table 3 and 5, our method generally performs skin detection with appropriate accuracy on very difficult and inaccurate semi-automatic ground truth from the COMPAQ database.

Table 5: Performances of Different Skin Detection Algorithms on The COMPAQ Dataset (Unit:%).

| Method | AUC | 1-EER |
|---|---|---|
| Jones and Rehg [33] | 94.2 | 88.0 |
| Sun [36] | 95.2 ≈ | 88.٨ ≈ |
| FCN8S (Zuo et al. [18]) | 94.78 | 88.21 |
| FCN8S+RNN (Zuo et al. [18]) | 95.93 | 90.18 |
| proposed method (N_New_ANFIS) | 92.99 | 91.05 |

≈ These values are computed from the ROC curves listed in corresponding Paper and not available in the original papers. Results for Jones and Rehg [33] and Sun [36] are taken from Zuo et al. [18].

In order to compare proposed method with other methods on the Pratheepan dataset , we calculated recall (R), precision (P), F-measure (F), false positive rate (FPR), false negative rate (FNR), true negative rate (TNR), total detection error (TDE), and accuracy (Acc) for evaluation protocols. The formulas for these protocols are given in Eqs. (8)- (15) ( Arsalan et al. [37], Kawulok et al. [38], Roy et al. [39]:

$$R = \frac{tp}{tp+fn} \quad (8)$$

$$P = \frac{tp}{tp+fp} \quad (9)$$

$$F = \frac{2RP}{R+P} \quad (10)$$

$$FRP = \frac{fp}{tn+fp} \quad (11)$$

$$FNR = \frac{fn}{tp+fn} \quad (12)$$

$$TNR = \frac{tn}{tn+fp} \quad (13)$$

$$TDE = FPR + FNR \quad (14)$$

$$Acc = \frac{tp+tn}{tp+fn+fp+tn} \quad (15)$$

Where tp, fn, fp and tn are true positive, false negative, false positive, and true negative, respectively. Here, tp is a pixel listed as a skin pixel in the ground truth and predicted as a skin pixel by the network output. fn is a pixel listed as a skin pixel in the ground truth but predicted as a non-skin pixel by the network. fp is a non-skin pixel in the ground truth but predicted as a skin pixel by the network. tn is a pixel listed as a non- skin pixel and correctly predicted as a non-skin by the network.

We compare our proposed skin detection method in the Pratheepan database with several methods from previous advanced works, such as: Bayesian [33], DLSD [39], multi-colour space skin segmentation (MSSS) [40], statistical skin model (SSM) of hybrid colour space and adaptive skin detection using face location (ASDFL) [41], a fusion strategy to detect skin (FASD) [42], face and body-based skin classification (FBSC) [43], Luminance adapted skin detection (LASD) [44], Fast propagation-based skin detection (FPSD) [45], discriminative skin-presence features (DSPF) [46], Kim et.al [47], Osman et.al [48] and Arsalan et.al [49] (see Table 6).

Bayesian method [33] employs color information. In 2017, Roy et al. [39] thanks to the slow speed of region- based convolutional neural networks (RCNN), used a Faster- RCNN (FRCNN) for hand detection and to reduce the false positive rate but the problem with this method is slowness of processing speed of a patch-based CNN in the ROI using the sliding window. The MSSS [40] and SSM methods perform poorly in updating the skin model adaptively. SSM of hybrid colour space [41] extract skin in images by using a hybrid colour space strategy and has good performance. ASDFL method [41] increases the robustness of the skin classification. FASD method [42] contains a lot of non-skin pixels, due to the false detecting of face regions. This method is a mixture of dynamic threshold and Gaussian method. results of FBSC [43] reduces false positives of skin detection but sometimes has unsatisfactory performance because of unstable facial skin extracting. LASD [44] is a luminance adapted color space method which helps to optimize least square error. Discriminative skin-presence features (DSPFs) FPSD [45] and DSPF methods [46] are based on seed propagation over the graph representation of images. Kim et.al [47] proposed a network-in-network structure based on the inception module of GoogLeNet and can be used for the problems of reconstructing an image. Their method has Complex inception network. Tan et al. [42] have much non-skin pixels in outputs. Osman [48] proposed an improvement on the skin detection based on multi-colour space. Although in their method a single threshold cannot strongly promise correct detection with a skin-like background but reduced the false positive rate in the detection of skin area and has better performance than Tan et al. [42]. Arsalan et.al [49] used Outer residual skip network (OR-Skip-Net) for skin segmentation in non-ideal situations. This network transfer the features extracted at the initial layers to the layers responsible for generating the final segmentation map. Their method is used 39 images from the Pratheepan database for training and another 39 images for testing to perform skin detection and it has a accuracy of 96.78. According to Table 6, accuracy of our proposed method is less than Arsalan et.al [49] and several methods in this Table, but in spite of the fact that the number of our training pixels is much less than Arsalan et.al [49], and the number of our test images is equal to Arsalan et.al [49] (39 images), the accuracy of our method is reasonable. According to Table 4, ROC curve and RMSE of proposed skin detection method (see Fig. 18 (b)), the efficiency of our method is appropriate. The test time of each image is between 3-4 seconds.

The training and testing of proposed method was executed on a desktop computer with an Intel® Core™ i7- 4500U CPU@ 1.80 GHz (4 cores), 8 GB RAM. In our method we do not use any pretrained models, such as ResNet, GoogleNet or etc. We designed our own network using MATLAB, R2016b.

Table 6: Comparison of our proposed method with existing methods on the Pratheepan dataset (unit:%).

| Method | Pratheepan dataset | | | | | | |
|---|---|---|---|---|---|---|---|
| | P | R | F | Acc | FPR | FNR | TNR |
| MSSS ( Mohanty and Raghunadh, [40] ) | 56.11 | 67.91 | 61.45 | 85.71 | n/a | n/a | n/a |
| SSM ( Luo and Guan, [41] ) | 69.23 | 74.92 | 71.96 | 90.21 | n/a | n/a | n/a |
| FASD ( Tan et al., [42] ) | 64.03 | 65.80 | 64.90 | 90.39 | n/a | n/a | n/a |
| FBSC ( Bianco et. al, [43]) | 74.70 | 75.04 | 74.87 | 91.55 | n/a | n/a | n/a |
| ASDFL ( Luo and Guan, [41] ) | 77.12 | 80.53 | 78.79 | 92.90 | n/a | n/a | n/a |
| Bayesian ( Jones and Rehg, [33] ) | 68.81 | 89.72 | 77.88 | 82.37 | n/a | n/a | n/a |
| LASD ( Hwang et al., [44]) | 79.54 | 82.75 | 81.11 | 83.61 | n/a | n/a | n/a |
| FPSD FPSS ( kawulok et al., [45] ) | 78.37 | 89.91 | 80.70 | 84.19 | n/a | n/a | n/a |
| DSPF ( M. Kawulok et al., [46] ) | 75.43 | 84.36 | 79.64 | 85.21 | n/a | n/a | n/a |
| CNIN gray ( Kim et al., [47] ) | 82.96 | 83.79 | 83.37 | 92.11 | n/a | n/a | n/a |
| CNIN Kim et al., [47]) | 90.03 | 89.12 | 89.57 | 94.83 | n/a | n/a | n/a |
| YCbCr-SV ( Osman et al., [48] ) | 91.49 | n/a | n/a | 84.05 | 6.9887 | n/a | n/a |
| CbCr-SV ( Osman et al., 48] ) | 91.48 | n/a | n/a | 83.48 | 6.9984 | n/a | n/a |
| YCbCr-RGB ( Osman et al., [48] ) | 90.04 | n/a | n/a | 83.48 | 8.3235 | n/a | n/a |
| CbCr-RGB ( Osman et al., [48] ) [ | 89.67 | n/a | n/a | 83.81 | 8.8040 | n/a | n/a |
| CbCr ( Osman et al., [48]) | 85.45 | n/a | n/a | 85.86 | 14.8803 | n/a | n/a |
| RGB ( Osman et al., [48] ) | 85.33 | n/a | n/a | 83.32 | 17.6375 | n/a | n/a |
| HSV ( Osman et al., [48] ) | 81.27 | n/a | n/a | 81.57 | 19.6069 | n/a | n/a |
| DLSD ( Roy et al., [39]) | n/a | n/a | 67.00 | 84.00 | n/a | n/a | 84.00 |
| OR-Skip-Net Arsalan et.al [49]) | 93.20 | 87.73 | 89.87 | 96.78 | 1.464 | n/a | 98.54 |
| proposed method (N_New_ANFIS) | ٨٧,٢٧ | ٩١,٤ | ٨٨,٩٩ | ٩٢,٩٣ | ٧,١١ | 8.60 | ٩٢,٨٨ |

∗Results for MSSS ( Mohanty and Raghunadh, 2016 [40]) , SSM ( Luo and Guan, 2017 [41] ), FASD ( Tan et al., 2012 [42] ), and FBSC ( Bianco et al., 2015 [43]) are taken from research on ASDFL ( Luo and Guan, 2017 [41] ). ∗∗Results for Bayesian ( Jones and Rehg, 2002 [33] ), LASD ( Hwang et al., 2013 [44] ), FPSD ( kawulok et al., 2013 [45]), DSPF ( M. Kawulok et al., 2014 [46] ), and CNIN gray ( Kim et al., 2017 [47]) are taken from research on CNIN Kim et al., (2017) [47] .

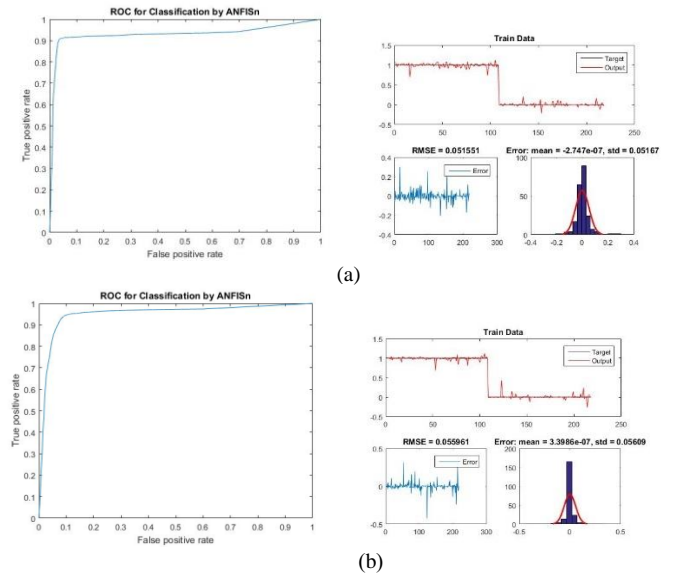

Figure 18. ROC curves of skin detection and RMSE values on the COMPAQ (a) and Pratheepan (b) datasets.

9. Conclusion

Color is an efficient feature for object detection as it has the advantage of being invariant to changes in scaling, rotation, and partial occlusion. This paper has highlighted the importance of skin detection by creating a new color space and comparing the result with the performance of RGB and LAB color spaces specifically in skin detection application. The test results of the proposed method (skin detection in new color space by nonlinear transformation using ANFIS classifier) in COMPAQ and Pratheepan databases showed that the proposed method is comparable to the advanced methods of skin detection due to its accuracy and speed in these two databases. This shows that creating new and optimized color spaces is useful for different applications in image processing and segmentation.

This paper has chosen a few pixels wisely for training BAO, which has performed skin detection accurately and quickly. However, one can increase the accuracy by selecting pixels of images with more varied races and colors intensities from the same database or other databases. However, in addition to accuracy, considering the algorithm speed optimized color spaces can be generated using other optimization algorithms or combined with other algorithms. For future works, deep learning of neural networks or other different types of membership functions in ANFIS can be useful for skin detection. Also, heuristic algorithms can be utilized to enhance ANFIS parameters.

Acknowledgment
Authors would like to thank Alimohammad Latif, Associated Professor, Computer Engineering Department, Yazd University, Yazd, Iran for his insights and helps.